\documentclass[conference]{IEEEtran}
\IEEEoverridecommandlockouts
\usepackage[T1]{fontenc}
\usepackage{cite}
\usepackage{amsmath,amssymb,amsfonts}
\usepackage{algorithmic}
\usepackage{graphicx}
\usepackage{textcomp}
\usepackage{xcolor}
\usepackage{subcaption, booktabs, microtype}
\begin{document}
\title{A Mobility-Aware Deep Learning Model for \\ Long-Term COVID-19 Pandemic Prediction and Policy Impact Analysis}
%
%
\author{\IEEEauthorblockN{1\textsuperscript{st} Danfeng Guo, Zijie Huang, Junheng Hao, Yizhou Sun, Wei Wang, Demetri Terzopoulos}
\IEEEauthorblockA{\textit{Department of Computer Science} \\
\textit{University of California, Los Angeles}\\
Los Angeles, USA}}

\maketitle              
\begin{abstract}
Pandemic(epidemic) modeling, aiming at disease spreading analysis, has always been a popular research topic especially following the outbreak of COVID-19 in 2019. 
Some representative models including SIR-based deep learning prediction models have shown satisfactory performance. However, one major drawback for them is that they fall short in their long-term predictive ability. Although graph convolutional networks (GCN) also perform well, their edge representations do not contain complete information and it can lead to biases. Another drawback is that they usually use input features which they are unable to predict. Hence, those models are unable to predict further future. We propose a model that can propagate predictions further into the future and it has better edge representations. In particular, we model the pandemic as a spatial-temporal graph whose edges represent the transition of infections and are learned by our model. We use a two-stream framework that contains GCN and recursive structures (GRU) with an attention mechanism. 
Our model enables mobility analysis that provides an effective toolbox for public health researchers and policy makers to predict how different lock-down strategies that actively control mobility can influence the spread of pandemics. Experiments show that our model outperforms others in its long-term predictive power. Moreover, we simulate the effects of certain policies and predict their impacts on infection control.

\end{abstract}
\section{Introduction}

COVID-19 is an infectious disease caused by SARS-CoV-2. This disease has rapidly spread all over the world since its discovery in December 2019. Recently, the wide spread of its Omicron variant has further exacerbated the pandemic and brought it back into the headlines. To date, millions of people have contracted COVID-19 and it has caused high death tolls, dramatically evolving from a healthcare system crisis into a worldwide socioeconomic disruption. This sharply increases the demand for pandemic models that predict the spread of infections.

Pandemic modeling refers to predicting disease-related statistics in a region (e.g., state-level daily new cases and deaths) based on historical data. It helps identify potentially high-risk groups, so that our public health officials can deploy medical resources in preparation for large-scale outbreaks of infection. Models that can simulate the effects of multiple alternative policies in order to assist the decision-making processes of policy makers would also be helpful.

\begin{figure}
    \centering
    \includegraphics[width=\linewidth]{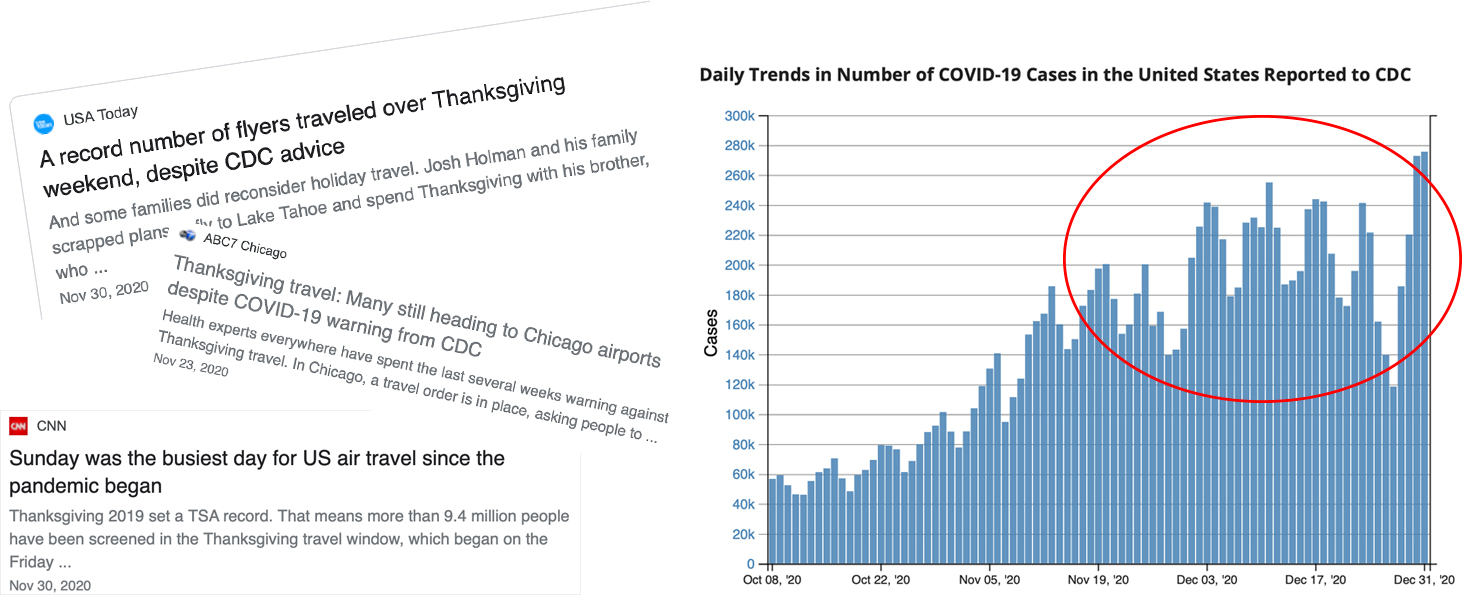}
    \caption{The explosion of new COVID-19 cases across the USA after Thanksgiving is relatable to high mobility nationwide during the holiday season.}
    \label{fig:mob_demo}
\end{figure}

The challenge of pandemic modeling lies not only in the availability of data, but also in the complexity of the available data. In addition to the pandemic statistics, one should also consider mobility data, which reflects the activity level of residents within a region or across regions. Mobility data provides information on the spread of disease, because higher mobility usually leads to more contact, which increases the chance of infection (Fig.~\ref{fig:mob_demo}). Other demographic and geographic statistics should also be included in pandemic modeling, such as the population. Since pandemic models may be used to assist decision making for public health policy, another challenge is that they are expected to predict further into the future---e.g., 3 weeks, one month, or even longer, rather than merely predicting the result tomorrow or next week. 

The majority of pandemic models extend the traditional SIR model \cite{sir}. It models the relation among susceptible group, infected group, and recovered group, and uses the transition rate to compute the number of newly infected people. Although they may be simple and sometimes less accurate, SIR-based models are reasonably predictive. In recent years, deep-learning models powered by data science and neural networks have been widely explored to predict pandemic growth. With regions represented as nodes in graphs, pandemic modeling is formulated as a graph mining problem. Graphical Convolutional Networks (GCN) \cite{KipfWelling} is an intuitive method to forecast pandemic growth based on regional and connectivity information. Recursive structures, such as the Gated Recurrent Unit (GRU) \cite{gru} and Long-Short Term Memory (LSTM) \cite{lstm}, have also been applied to encode pandemic progression through time. Recent GCN approaches have achieved success in spatio-temporal pandemic forecasting \cite{Cola}.

Compared with simpler SIR-based models, deep-learning-based models take a wider range of input features into consideration, are more sophisticated, and sometimes have better performance. However, they have certain explainability and interpretability issues. The representation of graph edges usually does not contain complete information. Another issues is that currently most deep-learning-based models are short-term models that cannot propagate their predictions into the long-term future. For example, they use all historical data only to predict the results on the 7th day from the present. If they must predict another day into the future, the models must be re-trained. Such models may have good short-term performance but are unsatisfactory in the long run. Most of them suffer from large error increases. In addition, they usually use some input features which they do not predict. The lack of those features prevents them from predicting further future. However, longer-term prediction results are much more helpful because they give our public health officials more time to respond. 

The deficiencies of existing epidemic models motivate us to develop a model that is able to propagate its predictions further into the future and has better edge design. In this paper, we propose a new two-stream, multitask GCN-GRU framework. It is a spatio-temporal model that encodes information on the graphical level during a time span. Our model exploits both pandemic statistics and mobility information. It has the following four main merits:
\begin{itemize}
    \item All input features we use is either static or can be predicted by our model. Hence, our model is able to propagate the predictions further into the future. It is also more robust in predicting long-term results. This is shown by comparing our model with others in terms of longer-term predictions.
    \item The edge weights in our model are learned from multiple factors which are related to pandemic spread. 
    \item Our model applies an attention mechanism which makes it focus on days most relevant to prediction.
    \item We are able to simulate our model under different mobility control policies and evaluate their impacts, which provides an insight to policy makers.
\end{itemize}

We test our model on the COVID-19 pandemic prediction task at the state level within the U.S., showing that it generally outperforms others on predicting cases 2 weeks ahead and 3 weeks ahead. We also simulate different mobility control policies and use our model to evaluate their effect in reducing confirmed cases. Lastly, our model is not limited to COVID-19, but can be applied to other epidemics which is related to mobility.

\section{Related Work}
\label{sec:rw}
\begin{table*}
    \centering
    \begin{tabular}{l p{2.0cm} p{2.0cm} p{5.2cm} c}
        \toprule
        \textbf{Models} & \textbf{Year} &\textbf{Structure} & \textbf{Edge Representation} & \textbf{Long-term prediction} \\
        \midrule
        ColaGCN\cite{Cola} & 2020 & RNN-GCN & Geographic adjacency & No \\
        Kapoor\cite{Kapoor2020ExaminingCF} & 2020 & GCN & Mobility \& time & No \\
        ITCGN\cite{itcgn} & 2021 & GCN & Geographic distance \& time & Yes \\
        STAN\cite{STAN} & 2021 & GCN-RNN & Demographic \& geographic similarities & No \\
        Fritz\cite{DBLP:journals/corr/abs-2101-00661} & 2021 & GCN & Social connection & No \\
        UNITED\cite{united} & 2021 & GCN-LSTM & Mobility & No \\
        La Gatta\cite{dp_sir} & 2021 & GCN-LSTM & Mobility & No \\
        \midrule
        Ours  & & two-stream GCN-GRU & Learned by model  & Yes \\
        \bottomrule
    \end{tabular}
    \caption{GCN pandemic models.}
    \label{tab:modelcmp}
\end{table*}

\paragraph{Traditional SIR/SEIR-based models}

The first epidemiology model to be considered is the SIR model \cite{sir}. It consists of three variables: $S$ for the number of susceptible individuals, $I$ for the number of infected individuals, and $R$ for the number of recovered or immune individuals. The incremental cases are determined by $S$, $I$, the regional population, and the computed transition rate. Subsequently, the SEIR \cite{seir} model introduced a new variable $E$, which is the number of individuals exposed to the disease yet remain asymptomatic. The SuEIR model \cite{sueir} makes a further improvement by including individuals that are infected but untested. DELPHI \cite{DELPHI} is another improved SEIR-type model that includes the underdetection group and outside intervention. \cite{policyspt} developed a recent SIR-based model that includes the infections not caused by mobility. \cite{nature} introduces an additional mobility network into SEIR that improves performance. 

\paragraph{Graph-based neural network models}

Graph analysis powered by machine learning has been developing quickly. GCN combined with recurrent structure has become a widely used baseline model for spatial-temporal data analysis. \cite{adaptive} uses an adaptive learning module to learn the feature of each node as well as the adjacency matrices. The learned variables are fed into a GCN to encode the spatial features. Then GRU is used to encode the temporal features. \cite{1dconv1,1dconv2} also use GCN, while 1D-convloution is applied to encode the temporal features. Later, \cite{gcnrnnmix} moves one step further and it embeds GCN into GRU by replacing the matrix multiplications with graphical convolutions. \cite{spectral} applies Discrete Fourier Transform to convert data from time domain to frequency domain and then performs graphical convolutions. Inspired by those works, researchers have started to formulate epidemic prediction as a dynamic graph problem, where the nodes represent target areas with some associated features.  \cite{Cola} proposes ColaGNN, which uses a Recurrent Neural Network (RNN) to encode the temporal features of each node and a GCN to encode the spatial features of the whole graph. The edge weights are the addition of the geographical adjacency matrix and a locally-aware attention map is computed using the RNN output features. By contrast, \cite{Kapoor2020ExaminingCF} applies a large GCN to encode both spatial and temporal information. At the spatial level the graph edges encode mobility, and at the temporal level the edges represent binary connections to earlier days. IT-GCN \cite{itcgn} also includes spatial and temporal nodes as a large graph whose edges represent the geographical distance and time connection to earlier days. \cite{DBLP:journals/corr/abs-2101-00661} uses a GCN to encode the mobility data. The encoder output is concatenated with the epidemic features and passed to a neural network for prediction. Another GCN-RNN model, STAN \cite{STAN}, designs nodes as regions with both static and dynamic features, and the edges represent demographic and geographic similarities. An attention mechanism is incorporated into the GNN for interaction between neighbors. \cite{united} uses GCN-LSTM model and designs the edges as the mobility between locations. The model performs well in predicting the COVID-19 pandemic in England, Spain, Italy, and France. \cite{dp_sir} creates a two-stage model that uses GCN-LSTM to compute the key factors in SIR, such as the contact rate.

Table~\ref{tab:modelcmp} summarizes the aforecited deep-learning-based pandemic models. A problem common to all of them is that their edge representations may not provide a good approximation of pandemic spread. They formulate their edge weights only using one of the infection data, mobility data, and other static regional features, which only contains partial information. For example, most models use mobility as edge weights and they ignore the fact that the number of disease-carriers can be small even though the mobility is large. A more causal and reasonable approach would include at least both the mobility data, infection statistics, and other demographic/geographic features. A second problem is that all these models set an $N$-day window length for data and predict the results after 1 day, 1 week, or 2 weeks. For example, the model of \cite{united} is trained to predict the number of cases on the 7th future day. Most of these models are unable to propagate to more distant days because some input data is unavailable. For instance, medical claim data used in \cite{STAN}, Facebook human mobility and connections used in \cite{DBLP:journals/corr/abs-2101-00661, united}, and the radius of people gyration used in \cite{dp_sir}. In addition, although their short-term results seem satisfactory, an obvious error increase is observed as the prediction time span increases and they may fail to predict 21 days or one month ahead. Although IT-GCN is able to predict long future, it only inputs daily case numbers and its use of geographical distance as edge weights is also problematic. Another problem lies in the recurrent structures. In traditional RNN/LSTM, the closest token usually has the largest influence on the results and farther-away tokens may be discounted or forgotten. However, this may not be the case for epidemic modeling, since data from earlier days may also play an important role.


\paragraph{Other models}

Aside from the above two categories, there exist other pandemic models: e.g., Gaussian-based models \cite{gaussian}, Hawkes Process models \cite{hawkes1,hawkard}, and Graph ODE models \cite{zijie}.

\section{Methodology}

\subsection{Problem Formulation}

At time $t$, a graph is represented as $G^{(t)}=(\textit{V}, \varepsilon^{(t)})$, where $\textit{V}$ is the set of nodes (with nodal features) and $\varepsilon^{(t)}$ is the set of directed, weighted edges. In our application, the nodes represent the regions at a certain level of granularity (e.g., cities, counties, or states). For each pair of nodes, there is one forward edge and one backward edge indicating the spread of disease from one state to the other. They are represented by an adjacency matrix $A\in \Re^{N\times N}$, in which element $A_{ij}$ indicates infections from state $i$ to state $j$. Within a time span $K$, there is a set of graphs $\{G^{(1)}, G^{(2)}, G^{(3)}, \dots, G^{(K)}\}$. Hence, the input feature set is formed as $X\in \Re^{K\times N\times D}$, where $N$ is the number of nodes and $D$ is the node feature dimension. The mobility input is $W\in \Re^{K\times N\times N}$

Given the inputs $X\in \Re^{K\times N\times D}$ and $W\in \Re^{K\times N\times N}$, our task is to design a model that predicts the graph of Day $K+1$; specifically, node features $X^{(K+1)}\in \Re^{N\times D}$ and the mobility $W^{(K+1)}\in \Re^{N\times N}$.
This muti-task learning strategy also yields two main advantages. Firstly, learning $X^{(K+1)}$ and $W^{(K+1)}$ shares similar features, so learning them simultaneously can make the model focus on learning these shared features, thus reducing overfitting. Secondly, we can use the predicted outputs as new inputs to predict the results of subsequent days.

\subsection{Edge Representation}

As mentioned earlier, existing deep-learning-based pandemic models do not have a good edge representation. An edge should represent the spread of infection and it results from the interaction of multiple complex factors including pandemic statistics, mobility patterns, demographic features, etc. Hence, we design our edges as variables learned from those factors. The idea of using learned edges has already been shown helpful in \cite{adaptive} and \cite{gcnrnnmix}. \cite{adaptive} points out that, for traffic forecasting problem, simply using one feature such as node distance or node similarity as edges does not provide complete information, which may result in considerable biases. Hence, they prefer the graphs to be learned instead of pre-defined. \cite{gcnrnnmix} proposes that the graphs can be learned when they are unknown. In our case, the pandemic graph can also be regarded as unknown because we do not have the number of infected travellers and it can only be estimated from multiple factors.  

\subsection{Model}

\begin{figure*}
    \centering
    \includegraphics[width=0.85\textwidth]{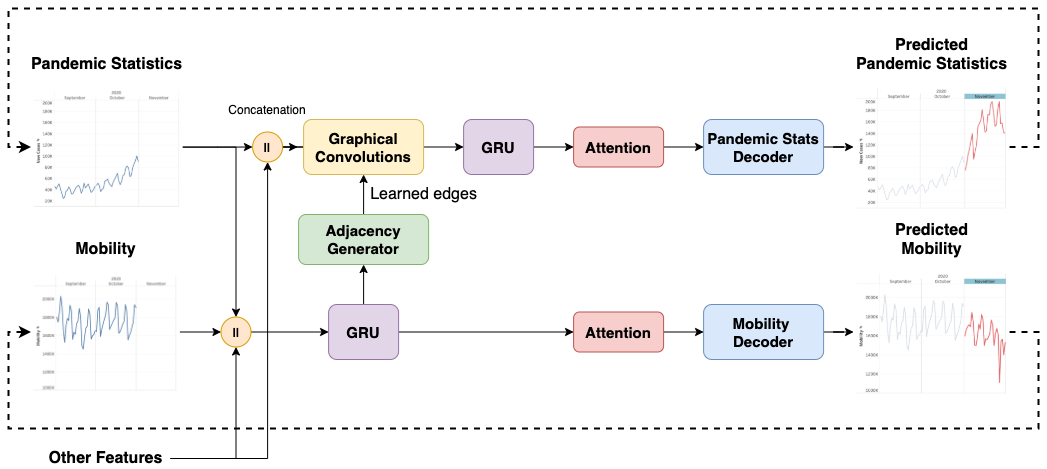}
    \caption{Our pandemic prediction model. Both predicted cases and mobility can be used as further inputs, enabling our model to predict the longer term.}
    \label{fig:model}
\end{figure*}

Given that we use mobility data as input, our model must be able to predict it so that it can propagate to the future. Hence, we use two-stream structure as shown in Fig.~\ref{fig:model}. The top stream takes pandemic statistics as inputs and returns the predicted pandemic statistics. It uses GCN to encode spatial features and GRU with attention to encode temporal features. The output of the GRU is fed to a Multi-Layer Perceptron (MLP) decoder to generate the predictions. The bottom stream uses previous mobility, pandemic statistics and demographic features to predict the future mobility and generate the edge weights for the GCN layers in the other stream. The inputs first come to a GRU encoder. The GRU output goes in two directions: one to an attention mechanism followed by an MLP decoder that generates the predicted mobility data, and the other goes to an MLP adjacency generator that generates the edges.

\paragraph{Learned Edges}
As mentioned previously, the bottom stream extracts information from previous mobility, pandemic statistics and demographic features. The extracted information comes to an adjacency generator, which acts as a multitask head to generate the edges at all time steps. The adjacency generator is a stack of linear blocks with LeakyReLU gates. Sigmoid activation is attached to the end.
\paragraph{GCN}

We use the GCN developed by \cite{KipfWelling}. For each node $v$, given its current representation $x_v$, neighbor $\mu \in \varepsilon$, and neighbor representation $x_\mu$, the GCN updates $x_v$ using its neighboring information. Generally,
\begin{equation}
    X_\text{new}=\sigma(\hat{D}^{-\frac{1}{2}}\hat{A}\hat{D}^{-\frac{1}{2}}XW),
\end{equation}
where $\hat{A}$ is the adjacency matrix with self-loops, $\hat{D}$ is a diagonal matrix where $D_{ii}=\sum_j A_{ij}$, the learnable parameter is $W$, and $\sigma()$ is the activation function. 

\begin{figure}
    \centering
    \includegraphics[width=0.95\linewidth]{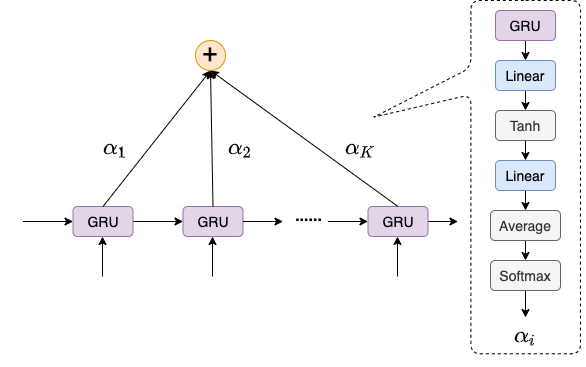}
    \caption{The GRU attention mechanism. The output is represented as a weighted sum of sequential features.}
    \label{fig:attention}
\end{figure}

\paragraph{GRU}

Our task involves the processing of sequential data, i.e., the dynamic changes of graph nodes and edges through time. Recurrent structures such as the GRU and LSTM are used to encode this type of information. Although the LSTM was a popular choice in prior work, we choose the GRU for our model. Both the GRU and LSTM aim to address the vanishing gradient problem. Compared with the GRU, the LSTM performs better in processing long sequences, but the sequence for our model will not be very long. Furthermore, the LSTM has more parameters and may easily overfit our limited amount of data. 

\paragraph{Attention Mechanism}

The attention mechanism of the GRU/LSTM was first developed by \cite{lstmattn}. The key idea is to use the hidden outputs of all tokens instead of only that of the last token. The final output is a Softmax weighted sum of the hidden outputs of all the tokens. The weight of each hidden output indicates the level to which the predicting target is related to the corresponding token features. In this way, the model can automatically search for tokens relevant to the predictions. The attention mechanism is formulated as
\begin{align}
    \begin{split}
        & \text{Output}=\sum_{t=1}^{K}\alpha^{(t)}h^{(t)},\\
        & \{\alpha^{(1)},...,\alpha^{(K)}\}=\text{Softmax}(g(h^{(1)}),\dots,g(h^{(K)}),
    \end{split}
\end{align}
where $h^{(t)}$ is the hidden output at position $t$ and $\alpha^{(t)}$ is the corresponding weight computed by passing $h^{(t)}$ to a neural network function $g()$. Fig.~\ref{fig:attention} shows the structure of our attention mechanism. To process the features at each time step, we use two linear layers with Tanh activation in between. Then the features at each time are averaged and passed to a Softmax function to become the feature weights. The final features are a weighted sum of all the features across the sequence. The use of the attention mechanism is important. Given that it usually takes days for an infected person to develop symptoms, we assume that the data of earlier days should have more importance than that of neighboring days. Additionally, when the model propagates the predictions to later days, previously predicted results are used as inputs to generate later results. In this situation, the attention mechanism can mitigate the influence of the previous prediction errors by assigning less weight.

\subsection{Training}

Our overall loss function is a multitask loss:
\begin{equation}
    L=w_1 \cdot \text{MAE}_{\text{case}} + w_2 \cdot \text{MSE}_{\text{case}} + w_3 \cdot \text{MSE}_{\text{mobility}},
\label{multiloss}
\end{equation}
where MAE and MSE stand for the Mean Absolute Error ($L_1$) loss and Mean Square Error ($L_2$) loss, respectively, and the hyperparameters $w_i$ balance the trade off between the two tasks.


\section{Experiments and Results}

\subsection{Experiment Setting}
We perform COVID-19 prediction at the state level based on data from the USA. The graph nodes represent the 50 states plus Washington, DC, for a total of 51 nodes. For our model inputs, we use the previous daily cases, cumulative cases and regional population to predict the upcoming daily cases. (population is fixed) Regarding the mobility prediction, we use previous mobility, cumulative cases and the weekday indicator (integers 0 through 6 represent Monday through Sunday) as inputs. We include weekdays because mobility has a weekly pattern that may be helpful to prediction. We choose the window length of previous data equal to 15. (We discuss this in our ablation study below)


Our mobility data comes from the SafeGraph Mobility dataset \cite{SafeGraph}, which contains rich mobility information at different granularities. Each location is assigned an ID and the movement from one ID to another is recorded. We collected mobility information from 2020/02/01 to 2021/03/31 (424 days) and grouped it by state. The COVID-19 statistics were obtained from the CDC Pandemic Data Tracker \cite{CDC}, which provides key statistics of COVID-19 for each state and DC. We extracted the state-level daily new cases from 2020/02/01 to 2021/03/31. The reason we choose this period is that, vaccinations in the U.S. started to increase rapidly since Apr 2021. The effect of vaccination is not included in our model. Therefore, data after Apr may lead to significant biases. Another reason is that, SafeGraph stopped providing daily mobility data after May 2021. The data from from 2020/02/01 to 2021/01/31 (366 days) were used for training and validation. The data from 2021/02/01 to 2021/03/31 (59 days) were used for testing. 


Our model was trained using the RMSProp Optimizer for 150 epochs. The learning rate is $10^{-3}$ and decays every 10 epochs by $10\%$. The weights in the multitask loss (\ref{multiloss}) were $w_1=1.0$, $w_2=1.0$, and $w_3=0.5$. Experiments were run on a Tesla P100. The results were evaluated using the Mean Absolute Error (MAE). Regarding the evaluation metrics, percentage error could be an option. However, we notice that the state-level cases vary hugely. The daily new cases of some states can be less than 10 while other states can have more than 30,000. Percentage error is not a suitable measure because a small error in states with few cases may lead to a very large percentage error. 

\subsection{Baseline Models}
We compare our model with widely-used benchmarks for state-level prediction of new cases in the US, including both SIR-based models and deep-learning-based models. They are:

\textbf{GCN-LSTM}: It is one of the most common spatial-temporal models. It is used by \cite{dp_sir,united}, a recent model for COVID-19 prediction and shares the most similarities with our model. It uses GCN to encode spatial information and LSTM to encode temporal information. The difference is that, it has one stream and the edge represents the mobility. We did not compare our model with those models directly because we lack specific pandemic statistics, such as the number of hospitalized patients in \cite{dp_sir}.

\textbf{GCN-TSFM}: As the success of transformers in natural language processing, people start to explore its potential on processing time-series data \cite{tsfmsurvey}. For traffic prediction tasks, the recurrent structure in GCN-RNN models could be replace by transformers, which can lead to performance improvement \cite{traffictsfm}. Hence, we also apply transformers to our model and report the performance. 

\textbf{ISOLAT-Mixture} \cite{Mixture}: Developed by MIT, it views the curve of daily cases as the sum of multiple Gaussian curves each of which represents an outbreak.

\textbf{Bucky} \cite{Bucky}: It is a spatially distributed SEIR model developed by Johns Hopkins University. It consists of a collection of multiple SEIR models, which are stratified via the age demographic structure of a geographic region. 

\textbf{SuEIR} \cite{sueir}: It is a variant of SEIR model proposed by UCLA. It includes the estimation of untested/unreported cases of COVID-19.

\textbf{DELPHI} \cite{DELPHI}: Similar to SuEIR, it includes underdetection and also considers the effect of government intervention. 

\begin{figure}
  \centering
  \subcaptionbox{2-weeks-ahead MAE for Feb 2021. Feb 6 is excluded because our model uses data up to Jan 31 for training. UCLA-SuEIR misses the prediction on Feb 27. \label{fig:50mob}}{\includegraphics[width=\linewidth]{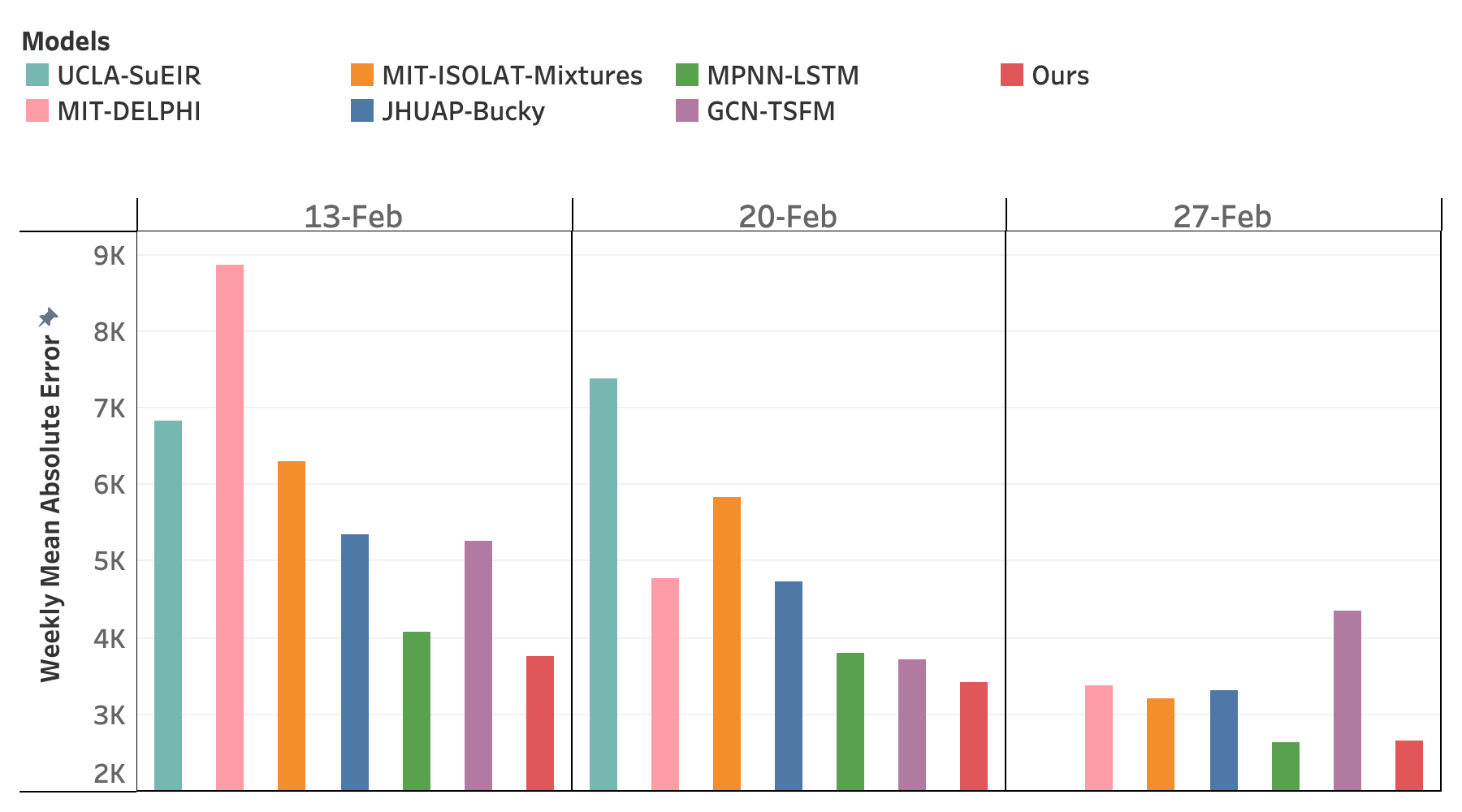}}\\[5pt]
  \subcaptionbox{2-weeks-ahead MAE for Mar 2021. \label{fig:interst}}{\includegraphics[width=\linewidth]{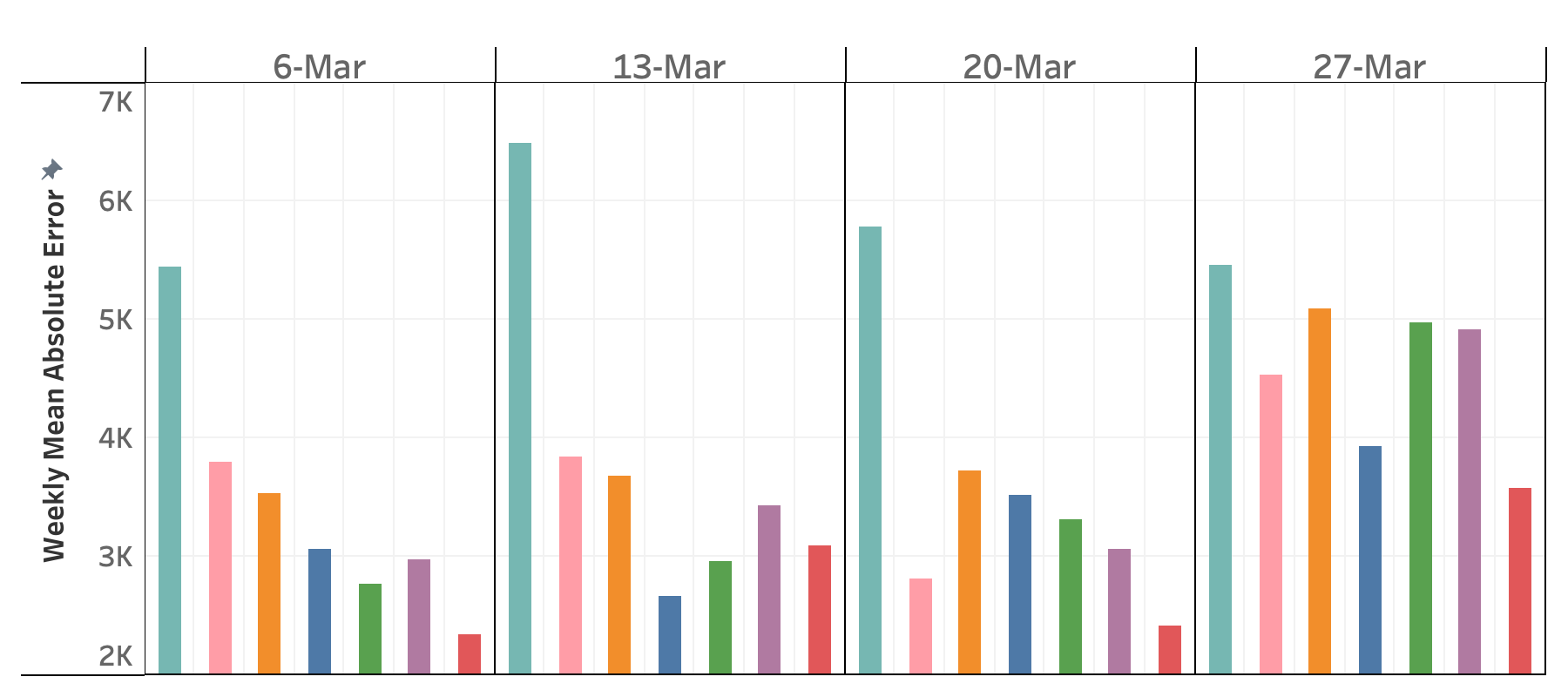}}\\[5pt]
  \subcaptionbox{3-weeks-ahead MAE for Mar 2021. UCLA-SuEIR misses Mar 6. \label{fig:isoca}}{\includegraphics[width=\linewidth]{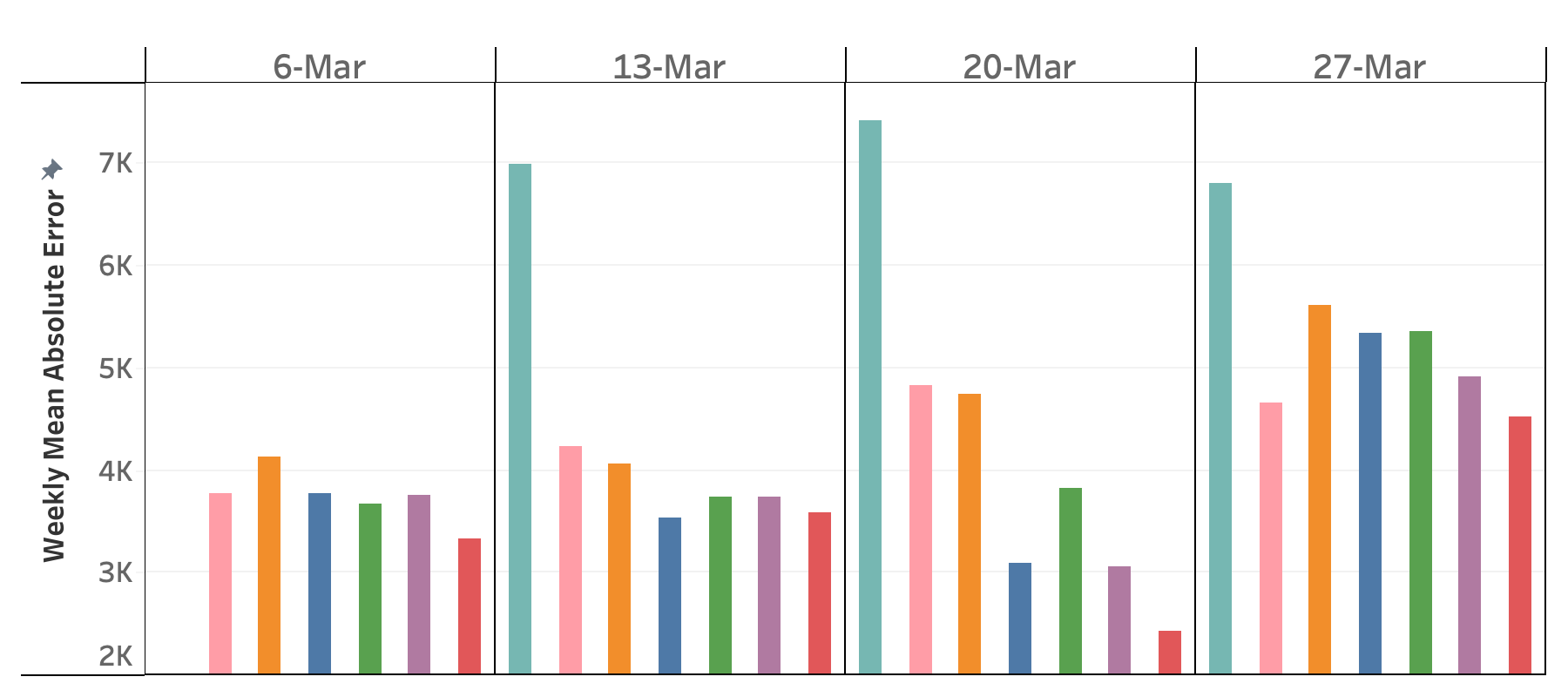}}\\[5pt]
  \caption{MAE comparison. The forecasts of ISOLAT-Mixture, Bucky, SuEIR and DELPHI were retrieved from the COVID-19 Forecast Hub repository \cite{cov19_forecast_hub}. }
  \label{fig:res_cmp}
\end{figure}

\subsection{Results}
The results are reported in the form of weekly cumulative new cases. The week used is the epidemiological week defined by the US CDC, which runs Sunday through Saturday. We made the comparison using results in February 2021 and March 2021. All predictions were made 2 weeks in the future, which means that only data up to 2 weeks prior to the target date (before Jan 17) were made available. We also generated predicted results using data up to 3 weeks prior, for which data before Jan 10 were available. The reason to choose 2 weeks and 3 weeks is to follow the rule of \cite{cov19_forecast_hub}, which is a public COVID-19 forecast repository where researchers report and compare their results. Predictions are all published as 2-weeks-ahead and 3-weeks-ahead.

The result comparisons are shown in Fig.~\ref{fig:res_cmp}. Note that our model has less error than GCN-LSTM, which illustrates that our learned edge representation works better than pure mobility. We also notice that transformer does not show expected performance on our task. One possible explanation could be the limitation of data. Transformers usually take a large amount of data to train, while we only have data for one year. Comparing across the two figures, one notices that several models suffer a large error increase when the prediction is changed to 3-weeks-ahead, as is summarized by Table~\ref{tab:tm_cmp}. Our model has the smaller error increase. GCN-TSFM has the minimum error increase, which benefits from its global attention mechanism. We belive that, in the future, with sufficient amount of data, it could achieve better performance.

\begin{table}
    \centering
    \begin{tabular}{l c}
    \toprule
        Models & Increase in MAE\\
        \midrule
        JHUAP-Bucky &  636\\
        MIT-ISOLAT Mixture & 640\\
        UCLA-SuEIR & 1279\\
        MIT-DELPHI & 630\\
        MPNN-LSTM & 650\\
        GCN-TSFM & 271\\
        \midrule
        Ours & 608 \\
    \bottomrule
    \end{tabular}
    \caption{Average MAE increase for March 2021 after changing the prediction from 2-weeks-ahead to 3-weeks-ahead.}
    \label{tab:tm_cmp}
\end{table}
\subsection{Ablation Studies}
\begin{figure}
    \centering
    \includegraphics[width=\linewidth]{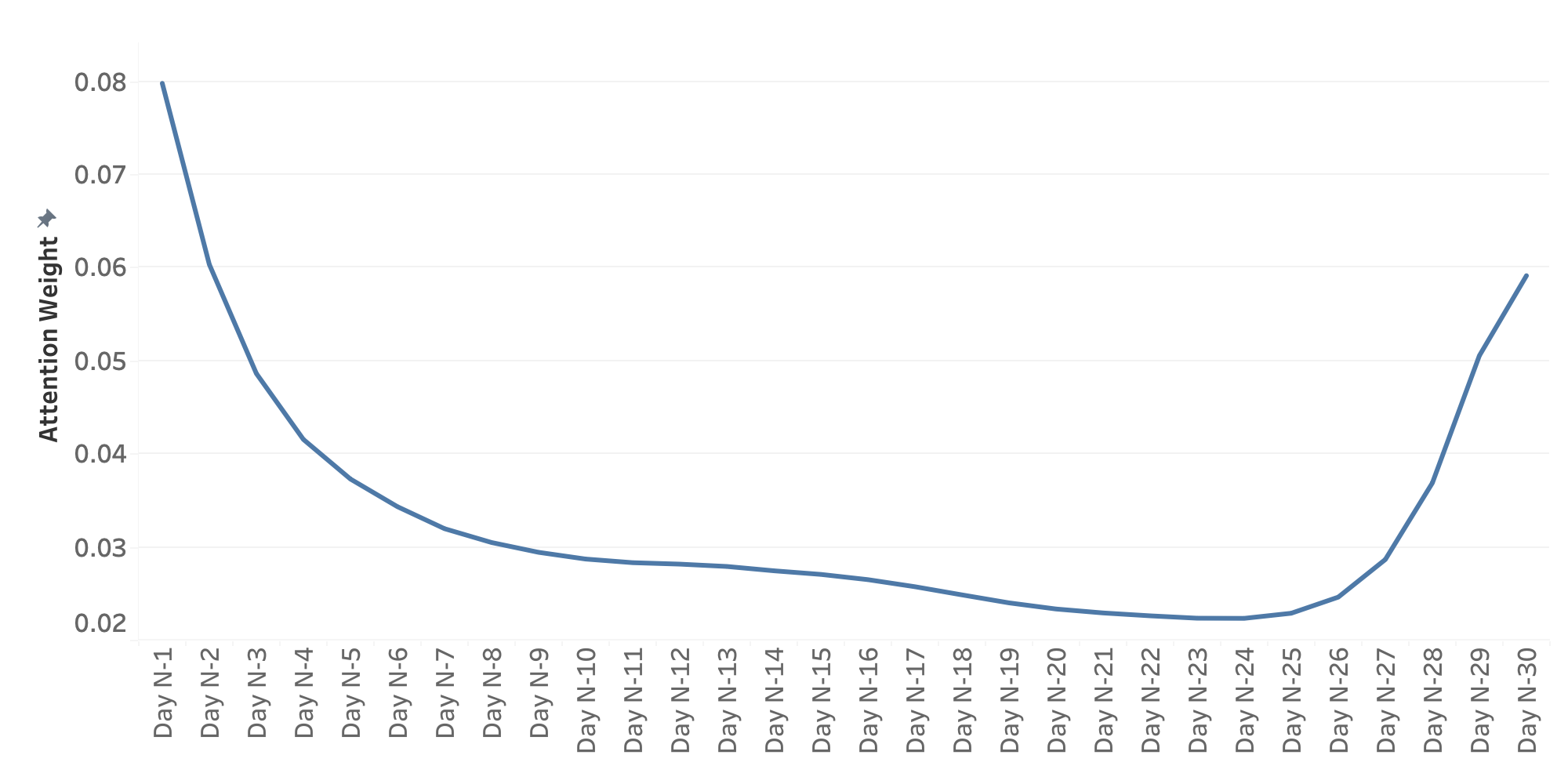}
    \caption{Average attention weights from Day $N-30$ to Day $N-1$. Length of data is 30.}
    \label{fig:attn_wgts}
\end{figure}

\begin{figure}
  \centering
  \subcaptionbox{ \label{fig:len_cmp}}{\includegraphics[width=0.95\linewidth]{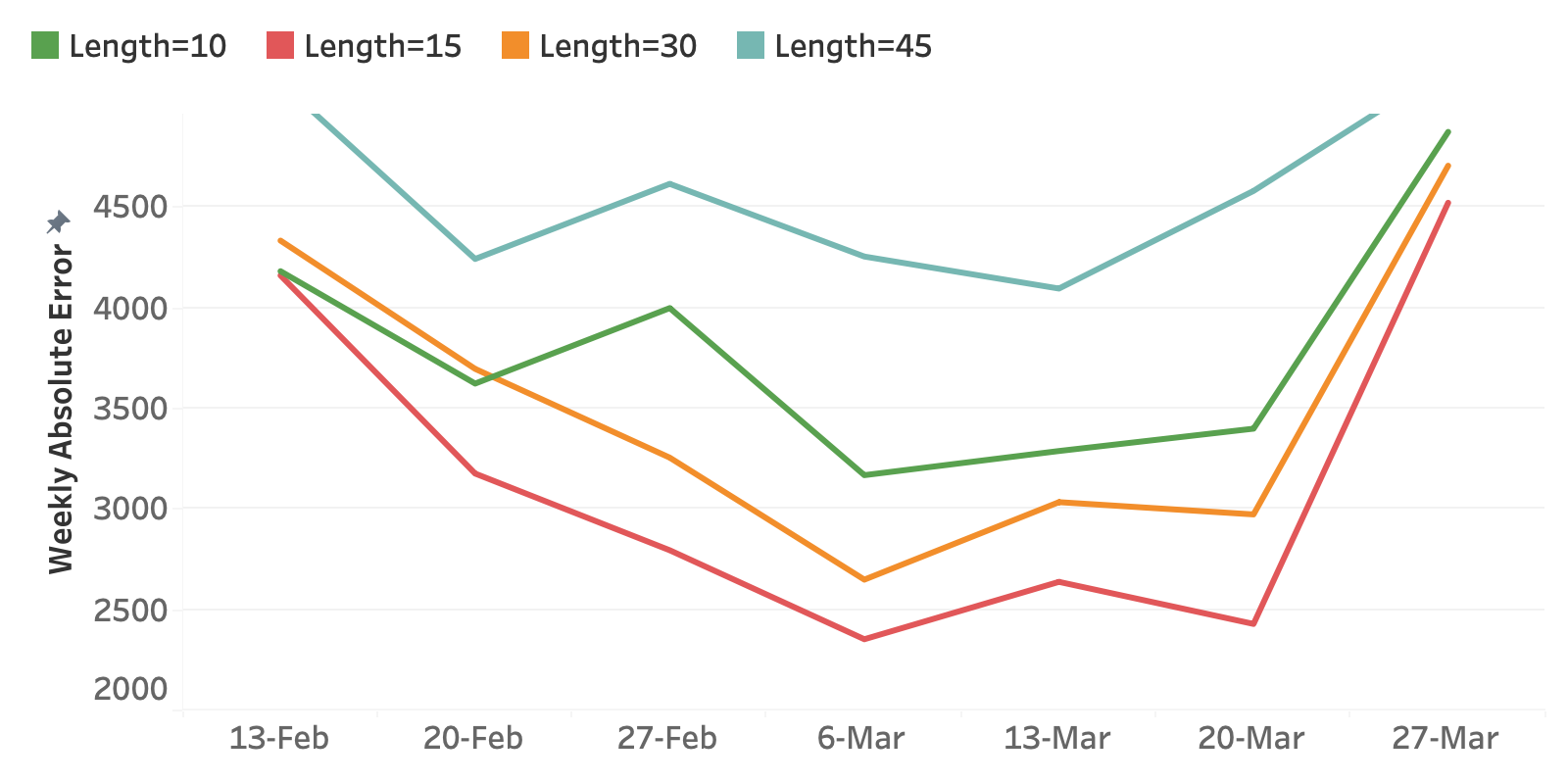}}
  \subcaptionbox{ \label{fig:attn_cmp}}{\includegraphics[width=0.95\linewidth]{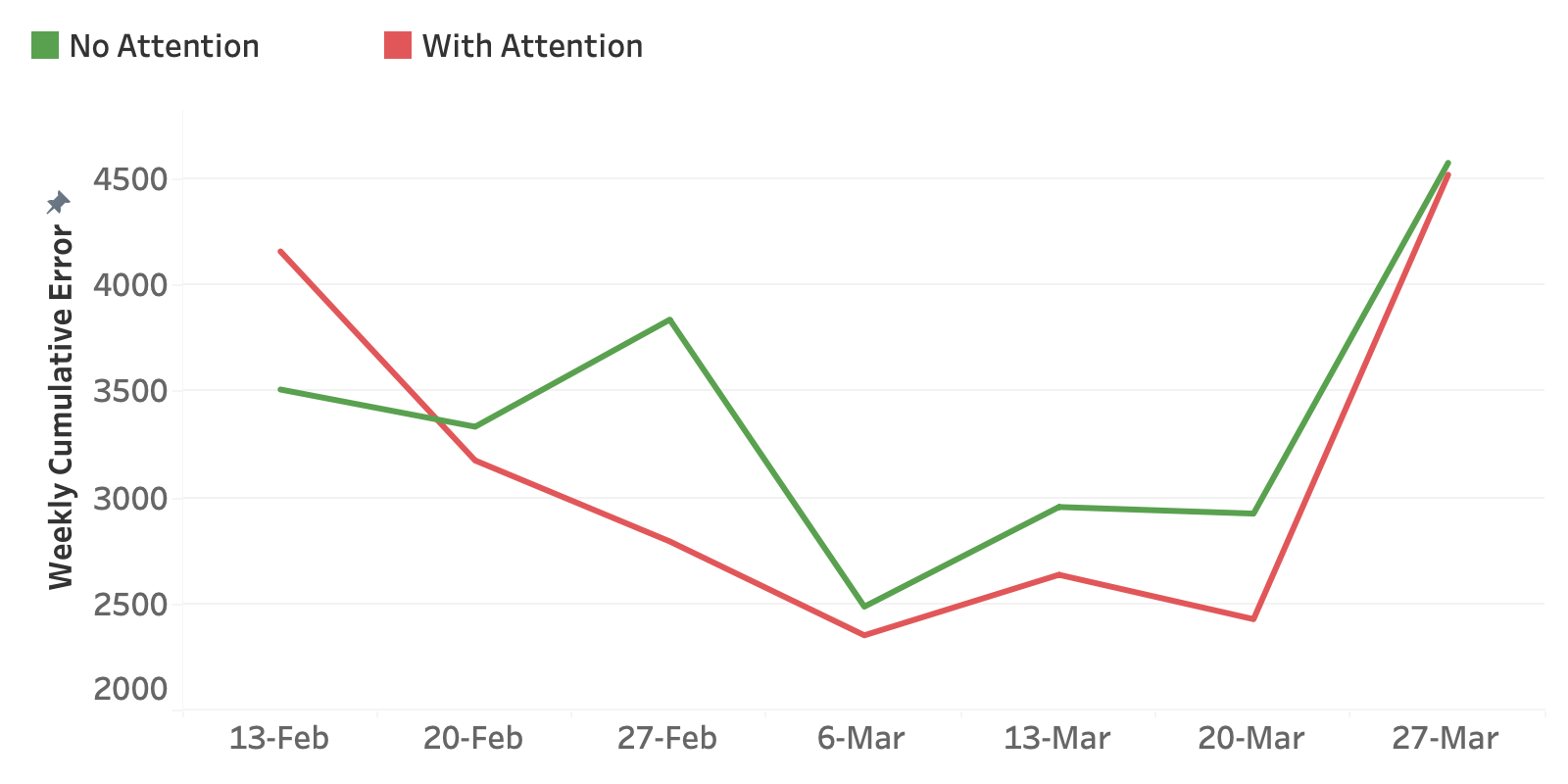}}\\
  \subcaptionbox{ \label{fig:w3cmp}}{\includegraphics[width=0.95\linewidth]{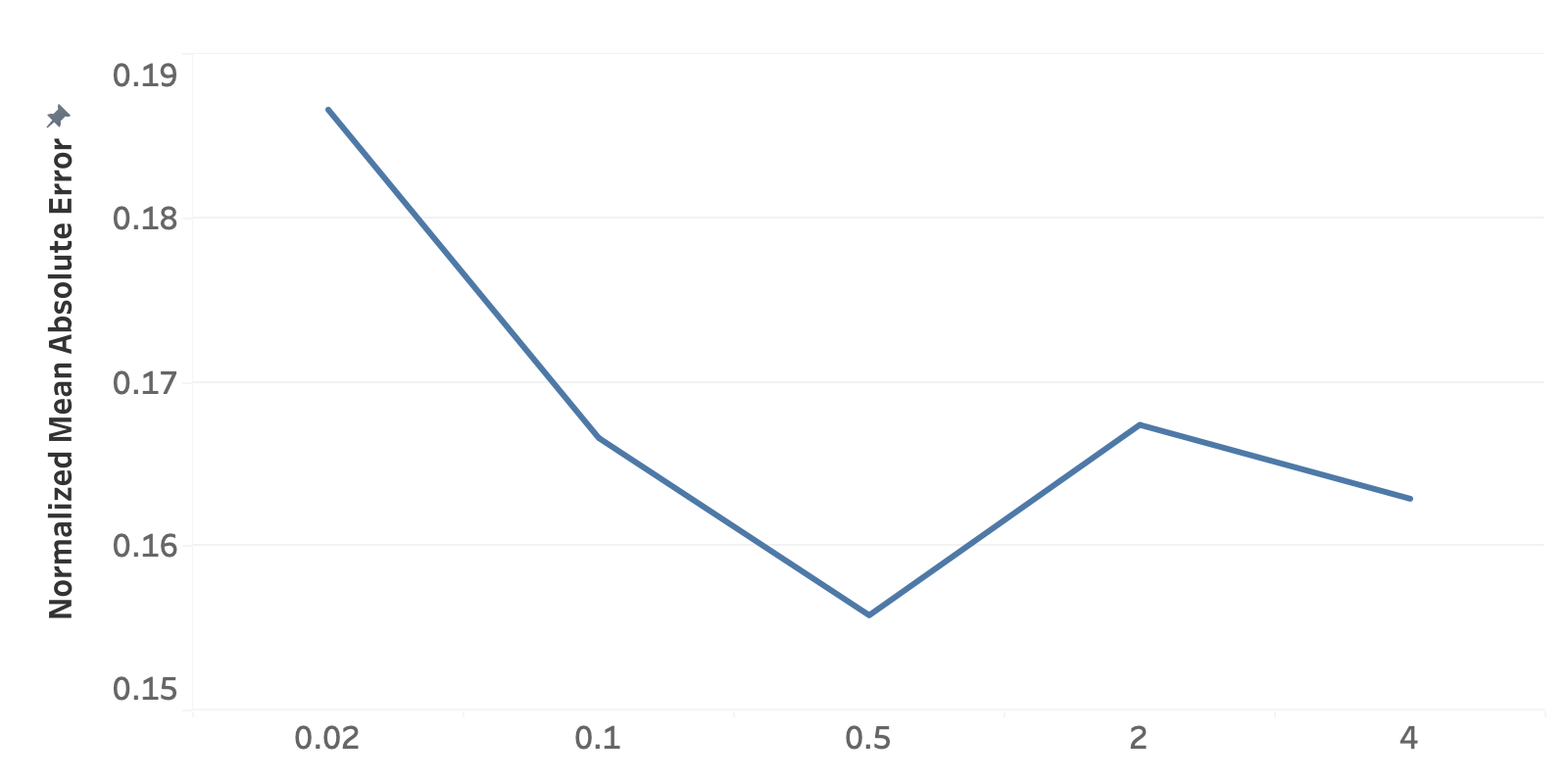}}\\  
  \caption{(a) Comparison of models with different time window lengths. (b) Ablation study of the attention mechanism. (c) Average MAE with different values of $w_3$.}
\end{figure}

The window length (number of days used for predictions) is an important parameter during the training phase. The attention mechanism provides guidance on tuning because the attention weights reflect how important one day's data is. Hence, we initially set the window length to $K=30$, and the average attention weights are shown in Fig.~\ref{fig:attn_wgts}. There is a valley from Day $N-17$ to Day $N-27$, which indicates that the data from 27 days prior to 17 days prior is less useful for prediction. Hence, the model may have better generalizability by choosing smaller window lengths. We conducted another three experiments, setting window lengths to 10, 15 and 45. The results are compared in Fig.~\ref{fig:len_cmp}. We choose $K=15$ as our final setting. Lastly, we conducted another experiment in which we disabled the attention mechanism to assess its contribution. The result, in Fig.~\ref{fig:attn_cmp}, shows that adding attention generally improves model performance.



We also studied how the multitask weights in Eqn.~(\ref{multiloss}) affect accuracy. We set $w_1=w_2=1.0$ and tuned $w_3$ over a set of values. The result is shown in Fig.~\ref{fig:w3cmp}, and we chose $w_3=0.5$ as the preferred setting.


\subsection{Policy impact analysis:}
Government intervention can play an important role in controlling the spread of pandemics. Policy makers should make effective intervention policies and mobility control is a crucial part of such policies; for example, the stay-at-home orders enacted in many countries, states, counties, and cities. Since mobility control policy has a broad impact on society as a whole, it is important to estimate its effect in advance, such that policy makers can assess the trade-off between the negative socioeconomic impact and infection control.

To this end, we used our model to simulate the effect of two mobility policies: reducing mobility by half and stopping interstate travel. For the first simulation, we manually reduced the mobility data by a half and for the second, we added all interstate mobility back to the origin states and zeroed out the interstate mobility. We assumed that the interstate travelers would choose to travel in their own states. All mobility manipulations start from 2021/02/01 and the policy impact is evaluated from 2021/03/01 by the cumulative change of new cases. The results are shown in Fig.~\ref{fig:50mob} and \ref{fig:interst}. 
Fig.~\ref{fig:50mob} reveals that reducing mobility by half does help to reduce the infections on a certain level. All states have the new cases reduced except for California. The cases in Georgia, New Jersey, South Carolina, and Florida reduce by more than 100. A more interesting result is Fig.~\ref{fig:interst}. In some states, cutting off interstate travel does substantially reduce the new cases. Especially in Missouri, the number of new cases decreases by over 1,000, which shows that a large portion of new infections come from outside instead of from the local residents. However, several states show an increase in cases. This phenomenon is also explainable as there are virus carriers among the interstate travelers. When they are kept in their origin states, there are likely to be more infections in those states. We further explored this policy by isolating one state, California. Per Fig.~\ref{fig:isoca}, most states show a decrease and this sheds light on the infection output of California.


\begin{figure}
  \centering
  \subcaptionbox{Reducing mobility by half \label{fig:50mob}}{\includegraphics[width=\linewidth]{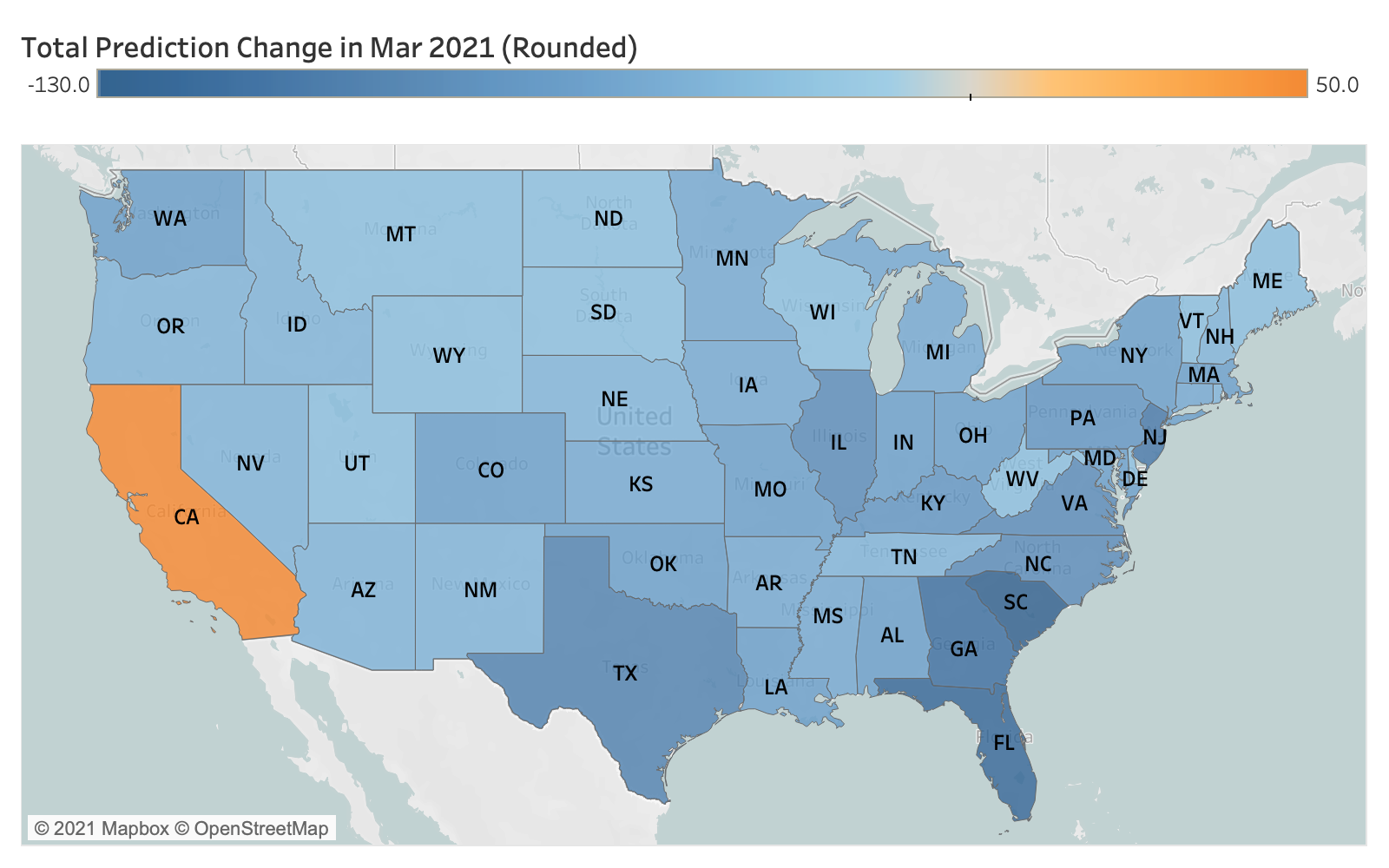}}\\[5pt]
  \subcaptionbox{Stopping interstate travel \label{fig:interst}}{\includegraphics[width=\linewidth]{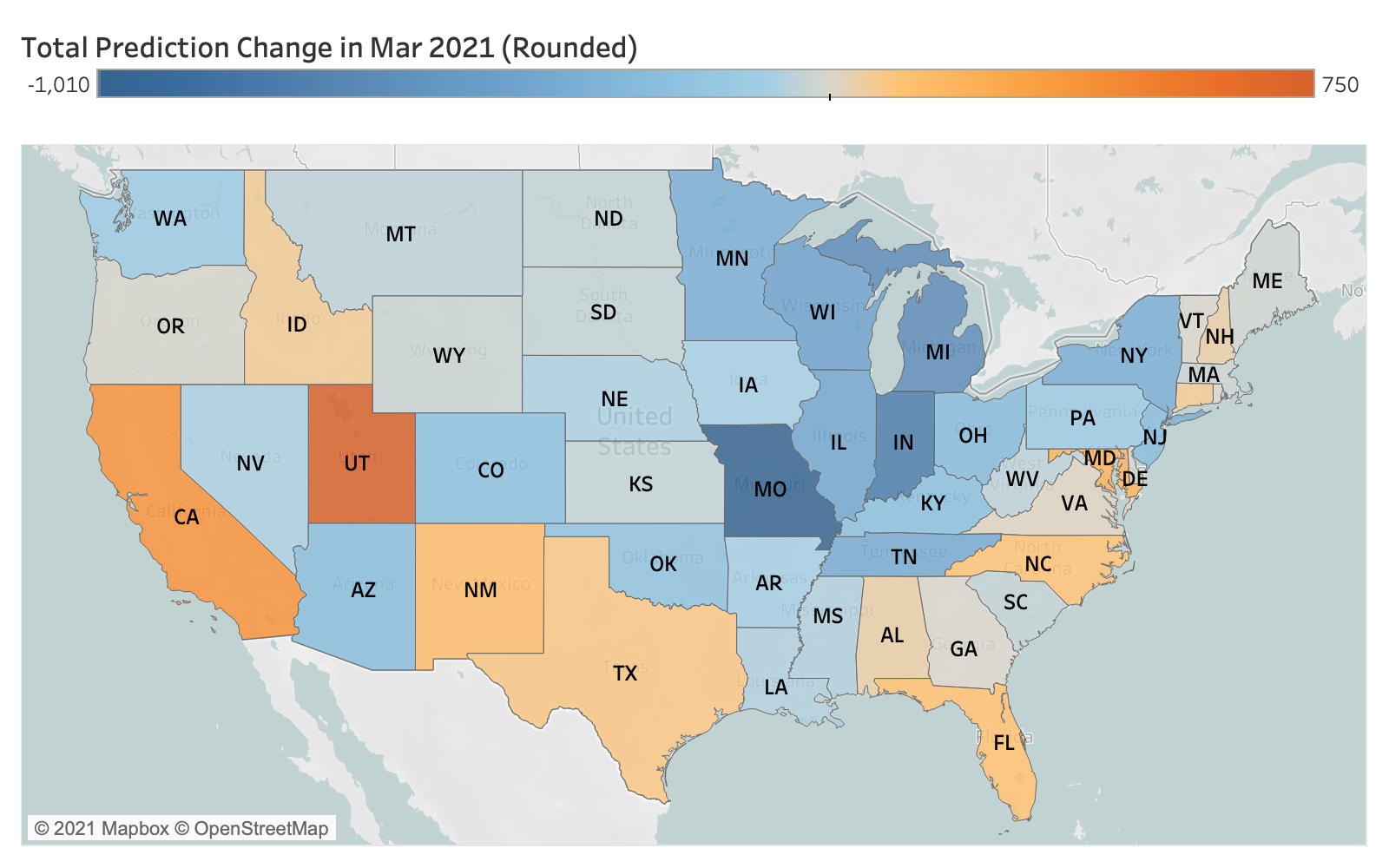}}\\[5pt]
  \subcaptionbox{Isolating California \label{fig:isoca}}{\includegraphics[width=\linewidth]{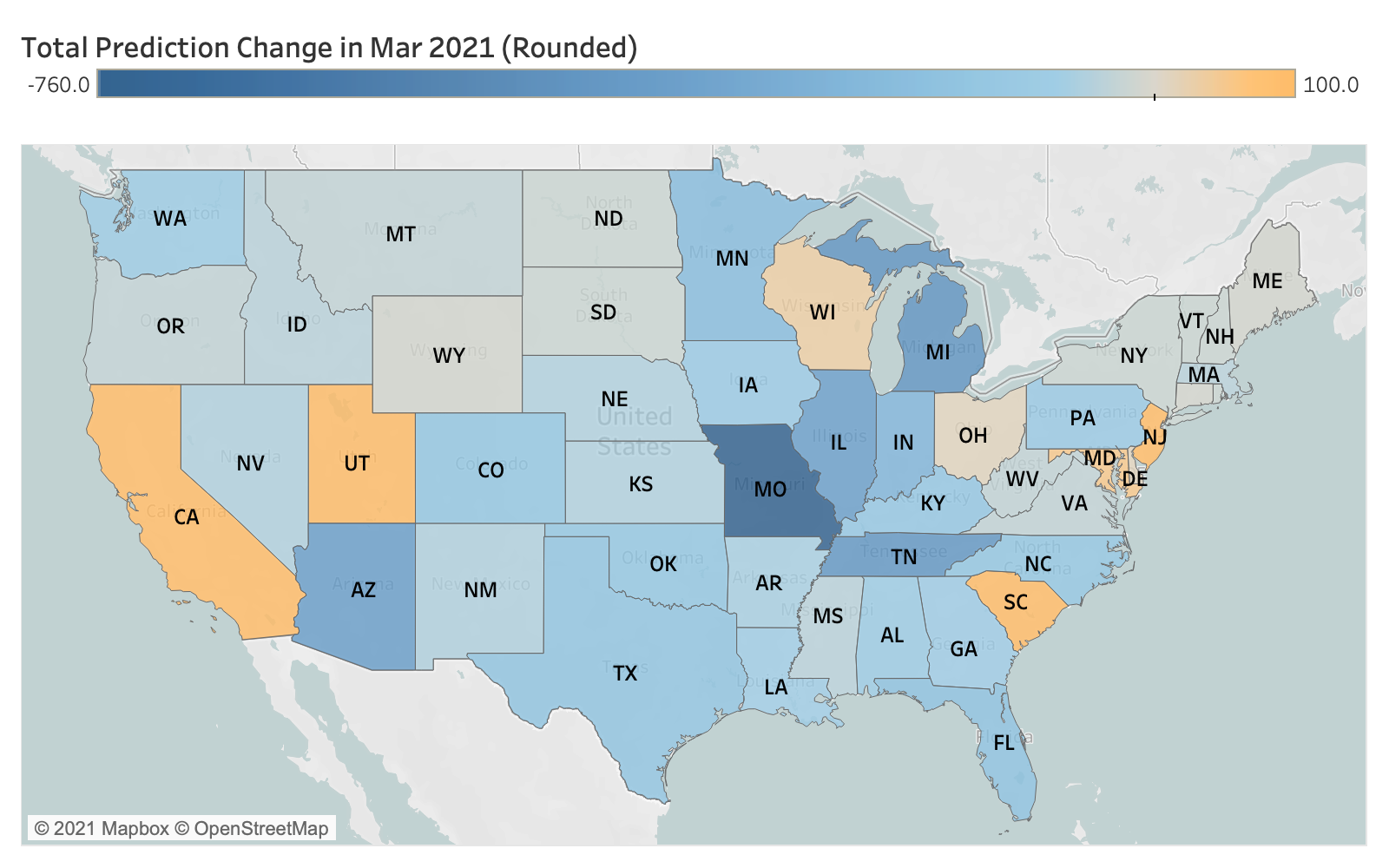}}\\[5pt]
  \caption{Change of total new case prediction in March 2021 with the policies of (a) reducing mobility by half, (b) stopping interstate travel, and (c) isolating California.}
\end{figure}

Neither Fig.~\ref{fig:50mob} nor \ref{fig:interst} show a large decrease in cases. One possible explanation could be that the USA experienced an infection peak in Feb, 2021, when the daily new cases nationally exceeded 100,000. Hence, those policies may not have a short-term influence, but could take longer to manifest their effect. 

We understand that validating our model using real policies would be more accurate. However, the effect of policies is hard to estimate because mobility control policies vary on smaller granularity level, e.g., counties. It is hard to estimate the overall effects on states. In addition, one can never know the ground truth cases where there are no such policies. Hence, the evaluation cannot be made.
\section{Conclusion and Discussion}

We have proposed a pandemic model that jointly predicts the daily new infection cases and the daily mobility and it is capable of predicting over the longer-term future. It is also able to simulate some mobility control policies and potentially assist in the decision-making process. Technically, the use of an attention mechanism improves model performance and provides insight on choosing the time window length. We are also the first to employ learned edges in GCN pandemic models and it outperforms GCN-LSTM that only uses mobility as edges. 

Modeling the effect of vaccination is a good future research direction as it is an important factor of pandemic models. Largely vaccinated groups may have fewer infections even with high mobility levels. We did not consider vaccination due to the unavailability of data. The number of vaccinations grew rapidly after March, 2021; however, we lack the corresponding data beyond that time. Moreover, there is no vaccination information in the training data. We may include the vaccination factor into our model when there are more data. 

There are sources of infection that our model is unable to cover because they lack representation in our mobility data; e.g., household infections and infections caused by mobility absent from the SafeGraph dataset. This could be a major source of prediction error, but it can easily be incorporated into SIR \cite{policyspt} and could be another direction for future work. In addition, mobility is not directly linked to infection. Direct/indirect contact is. Hence, relying on mobility data may reduce the confidence of our predictions. Lastly, more statistics can also be included in future work, such as the number of deaths and other demographic features. Our model can also be generalized to support different types of input.

Lastly, there are ethical concerns which should be addressed because the data involves privacy issues. Collecting mobility data poses specific challenges for upholding ethics principles. For the data we use, it was collected by SafeGraph after receiving the consent of the users. Strict procedures must be followed to protect people's privacy. On the other hand, given that some people may not be willing to provide their mobility patterns, the collected data may not be very representative. In this case, the accuracy is expected to drop. 

%
%
%
\bibliographystyle{IEEEtran}
\bibliography{ref}
%




\end{document}